\documentclass{IEEEoj}
\usepackage{cite}
\usepackage{amsmath,amssymb,amsfonts}
\usepackage{algorithmic}
\usepackage{optidef}
\usepackage{graphicx,color}
\usepackage{textcomp}
\usepackage{optidef} 
\usepackage{booktabs} 
\usepackage{tabularx} 
\usepackage{multirow} 
\usepackage{lipsum} 
\usepackage{bm} 

\newcommand{\<}{\mkern1mu} 
\def\BibTeX{{\rm B\kern-.05em{\sc i\kern-.025em b}\kern-.08em
    T\kern-.1667em\lower.7ex\hbox{E}\kern-.125emX}}
\AtBeginDocument{\definecolor{ojcolor}{cmyk}{0.93,0.59,0.15,0.02}}
\def\OJlogo{\vspace{-14pt}\includegraphics[height=28pt]{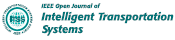}}
\begin{document}
\receiveddate{XX Month, XXXX}
\reviseddate{XX Month, XXXX}
\accepteddate{XX Month, XXXX}
\publisheddate{XX Month, XXXX}
\currentdate{XX Month, XXXX}
\doiinfo{OJITS.2022.1234567}

\title{Estimating the Joint Probability of Scenario Parameters with Gaussian Mixture Copula Models}

\author{CHRISTIAN REICHENBÄCHER\authorrefmark{1,2}, PHILIPP RANK\authorrefmark{1}, JOCHEN HIPP \authorrefmark{1}, AND OLIVER BRINGMANN \authorrefmark{2}, MEMBER, IEEE}
\affil{Mercedes-Benz AG, Sindelfingen, Germany}
\affil{Department of Computer Science, University of Tübingen, Tübingen, Germany}
\corresp{CORRESPONDING AUTHOR: Christian Reichenbächer (e-mail: christian.reichenbaecher@mercedes-benz.com).}
\markboth{Preparation of Papers for IEEE OPEN JOURNALS}{Author \textit{et al.}}

\begin{abstract}
This paper presents the first application of Gaussian Mixture Copula Models to the statistical modeling of driving scenarios for the safety validation of automated driving systems. Knowledge of the joint probability distribution of scenario parameters is essential for scenario-based safety assessment, where risk quantification depends on the likelihood of concrete parameter combinations. Gaussian Mixture Copula Models bring together the multimodal expressivity of Gaussian Mixture Models and the flexibility of copulas, enabling separate modeling of marginal distributions and dependence. We benchmark Gaussian Mixture Copula Models against previously proposed approaches—Gaussian Mixture Models and Gaussian Copula Models—using real-world driving data drawn from two scenarios defined in United Nations Regulation No. 157. Our evaluation on approximately 18 million instances of these two scenarios demonstrates that Gaussian Mixture Copula Models consistently surpass Gaussian Copula Models and perform competitively with Gaussian Mixture Models, as measured by both log-likelihood and Sinkhorn distance, with relative performance depending on the scenario. The results are promising for the adoption of Gaussian Mixture Copula Models as a statistical foundation for future scenario-based validation frameworks.
\end{abstract}



\maketitle
\section{INTRODUCTION}

\IEEEPARstart{S}{afety} of the Intended Functionality (SOTIF) has emerged as a critical aspect in the development of automated driving systems, focusing on minimizing risks arising not only from system failures but also from functional insufficiencies and performance limitations~\cite{noauthor_road_2022}. Unlike functional safety, which is governed by ISO 26262 and addresses hazards caused by malfunctioning behaviors of the electrical/electronic (E/E) system~\cite{noauthor_road_2018}, ISO 21448:2022 \textit{Road vehicles—Safety of the Intended Functionality} (SOTIF) emphasizes addressing hazards linked to the intended functionality itself, such as environmental perception errors or unexpected algorithmic behaviors \cite{noauthor_road_2022}.

\begin{figure}
\centering\includegraphics[width=1.0\columnwidth]{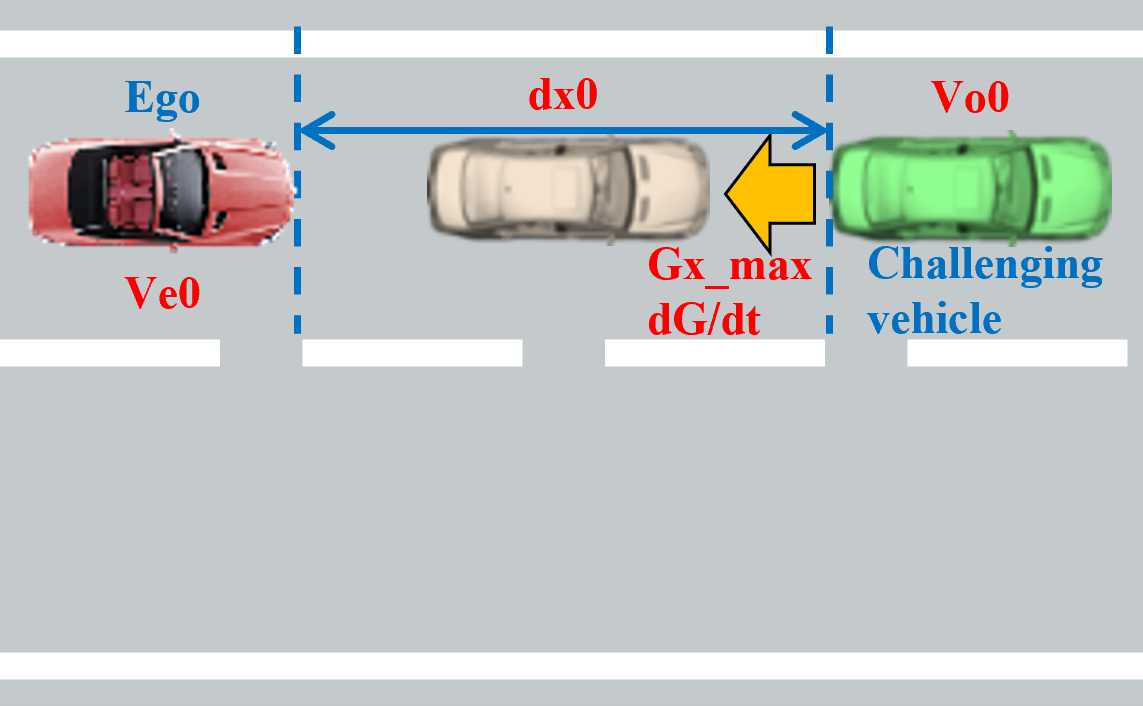}
\caption{Illustration of the \textit{Deceleration} scenario as defined in UN Regulation No. 157 (UN R157). For a detailed explanation of the scenario and the associated parameters, see Section \ref{sec:experiment}. Scenarios like this motivate the need for probabilistic modeling of parameter combinations in safety validation. Figure adapted from \cite{noauthor_uniform_2021}.}
\label{fig:deceleration}
\end{figure}

Achieving safety levels comparable to human drivers requires extensive testing and validation \cite{reichenbacher_identifying_2022}. However, given the impracticality of real-world testing—which would require billions of kilometers of driving data and substantial financial and time investments~\cite{maurer_millionen_2015}—simulation has become an indispensable tool for scenario-based assessments (see also~\cite{stadler_credibility_2022, genser_occlusion_2023}). Scenario-based approaches decompose the overall validation task into logical traffic scenarios, such as cut-ins or lead vehicle decelerations~\cite{menzel_scenarios_2018, mazzega_pegasus_2019, rasch_safety_2019}. These scenarios are defined by key parameters, such as braking deceleration and inter-vehicular distance, as illustrated in the \textit{Deceleration} scenario in Fig. \ref{fig:deceleration}. By systematically varying these parameters within simulation environments, the space of potential scenarios is explored to identify critical combinations that pose safety risks.

Central to this approach is the ability to determine the joint probability of specific parameter combinations, as this directly underpins the calculation of risk and ensures that simulated scenarios are relevant for real-world operation. ISO 21448 defines risk as the ``combination of the probability of occurrence of harm and the severity of that harm”~\cite[subclause~3.23]{noauthor_road_2022}. Without precise knowledge of these probabilities, the residual risk cannot be effectively quantified, nor can the most critical scenarios for validation be prioritized~\cite{most_multimodal_2023}.

The evaluation of joint probabilities for parameter combinations aligns directly with the requirements of ISO 21448 and UN R157 for robust risk assessment. ISO 21448 entails a systematic evaluation of residual risk for known and unknown hazardous scenarios, which requires not only identifying these scenarios but also quantifying their likelihood \cite{noauthor_road_2022, heintzel_statistical_2024}. Similarly, UN R157 mandates that Automated Lane Keeping Systems (ALKS) systems are validated under a broad range of operational conditions, ensuring the system’s robustness across diverse and plausible traffic scenarios \cite{noauthor_uniform_2021}. Both standards implicitly demand insights into real-world parameter distributions and dependencies, as these are the foundation for robust scenario-based validation processes like the one outlined by Most et al., which prioritize safety-critical scenarios through probabilistic analysis~\cite{most_multimodal_2023}.

The learning of joint probability distributions from a given set of data is referred to as generative modeling in the field of statistics and machine learning~\cite{gm_comprehensive_2020}. In general, the goal of generative modeling is to create new data points that resemble the distributions and properties of the original data~\cite{gm_comprehensive_2020}. In our case, however, we not only want to generate new data points, but also quantify the likelihood of a specific data point under the model distribution. Therefore, this is also referred to as generative modeling with explicit density estimation~\cite{gm_comprehensive_2020}.

Recent studies on the probabilistic representation of driving-scenario parameters therefore rely on two main model families.

\begin{itemize}
    \item \textbf{Gaussian Mixture Models (GMMs)} approximate the joint density with a weighted sum of multivariate Gaussian components~\cite{bishop_mixtures_2006} and have been successfully applied to driving scenarios, such as the highway cut-in risk analysis by Akagi et al.~\cite{akagi_risk-index_2019}. GMMs are well known to be universal approximators of continuous distributions given a sufficiently large number of components~\cite{bishop_gaussian_2006}. In practical applications, however, the number of components is typically finite, which reflects a trade-off between flexibility and robustness. In this finite-mixture regime, the elliptical structure of individual components may lead to oversmoothing of localized or strongly non-Gaussian features, particularly when complex marginal effects dominate.
    \item \textbf{Copula-based models}, such as the Gaussian copula used by Most et al.~\cite{most_multimodal_2023}, separate marginal and dependence modeling, enabling each variable to follow its own non-Gaussian distribution. However, a single elliptical copula enforces a globally monotonic dependence structure and cannot resolve multimodality or localized tail effects~\cite{aas_pair-copula_2009}. Vine (pair-copula) constructions push flexibility further by assembling many low-dimensional copulas into an R-vine, as demonstrated for urban roundabout data by Lotto et al.~\cite{lotto_modeling_2022}. These can capture complex, non-elliptical dependencies but, require a large number of pairwise copulas and become increasingly complex and difficult to interpret in higher dimensions, as noted by the authors.
\end{itemize}

In this paper, we explore \textbf{Gaussian Mixture Copula Models (GMCMs)} as an extension of mixture-based modeling that retains the dependence structure of GMMs while increasing flexibility through separate, potentially non-parametric, marginal modeling. Specifically, GMCMs use a Gaussian-mixture-based copula construction that decouples marginal and dependence modeling. While GMCMs have been proposed and analyzed in general machine learning contexts~\cite{tewari_parametric_2011, bhattacharya_unsupervised_2014, bilgrau_gmcm_2016, kasa_improved_2022, tewari_estimation_2023}, they have not yet been applied to the domain of automated driving.

This work presents the first application of GMCMs to the estimation of joint probability distributions of scenario parameters in driving scenarios. We demonstrate how GMCMs can be used to model complex, multimodal dependencies in safety-critical scenarios, as required by ISO 21448 and UN Regulation No. 157. Our evaluation is based on approximately 18 million real-world driving scenarios and includes a quantitative comparison with GMM and Gaussian Copula Models (GCM) using both log-likelihood and Sinkhorn distance metrics.

Our results establish GMCMs as a promising tool for scenario-based validation of automated driving systems, enabling more accurate risk quantification and prioritization of critical scenarios.

\section{THEORETICAL FOUNDATIONS}

\subsection{PROBABILITY DISTRIBUTION OF RANDOM VARIABLES}

Following the foundational work by \cite{tewari_parametric_2011}, we provide a brief recap of fundamental concepts regarding joint distributions of random variables, which serve as a basis for discussing GMCMs in the remainder of this work.

The joint cumulative distribution function (or simply distribution), $F: \mathbb{R}^d \to [0,1]$, for a set of $d$ real valued random variables, denoted as $X_1, X_2, \dots X_d$, is defined as
\begin{equation}
    F (x_1, x_2, \dots x_d; \Theta) = \mathrm{Pr} [X_1 \leq x_1, X_2 \leq x_2, \dots X_d \leq x_d],
\end{equation}
where $x_j$ is a realization of the variable $X_j$ and $\Theta$ denotes the parameter set characterizing the joint distribution.

The marginal distribution of a single variable $X_j$ is obtained by marginalizing over all other variables, which effectively reduces the joint distribution to its projection on dimension $j$ \cite{tewari_parametric_2011}:
\begin{align*}
    F_j (x_j) = \lim_{x_i \to \infty, \forall i \neq j} F (x_1, x_2, \dots x_d); \quad i, j = 1\dots d.
\end{align*}
Since the marginal distribution values are uniformly distributed in the interval $(0,1)$, an inverse transformation can be uniquely defined as
\begin{align*}
    F^{-1}_j (t : t \sim U(0,1)) \to \inf [x_j : F_j(x_j) \geq t]
\end{align*} 
and allows the transformation of a scalar between 0 and 1 to a unique value of the random variable $X_j$, following \cite{tewari_parametric_2011}.

Provided that the joint distribution $F(x_1, \dots x_d)$ is sufficiently differentiable, the corresponding joint probability density function (or simply density), $f: \mathbb{R}^d \to (0, \infty)$, is given by:
\begin{align*}
    f (x_1, x_2, \dots x_d) = \frac{\partial^d F (x_1, x_2, \dots x_d)}{\partial x_1 \partial x_2 \dots \partial x_d}.
\end{align*}

\subsection{GAUSSIAN MIXTURE MODEL}

A GMM describes the joint density of random variables as a weighted sum of $K$ normal distributions,  
\begin{align}
\label{eq:gmm}
    \psi (x_1, x_2, \dots x_d; \Theta) = \sum_{k=1}^{K} \alpha^{k} \, \phi(x_1, x_2, \dots x_d; \bm{\mu}^{k}, \Sigma^{k}),
\end{align} 
where $\alpha^{k} \geq 0$ denotes the weight of the $k$-th component, such that $\sum_{k=1}^{K} \alpha^{k} = 1$. $\phi(x_1, x_2, \dots x_d; \bm{\mu}^{k}, \Sigma^{k})$ represents a multivariate normal distribution with mean vector $\bm{\mu}^{k}$ and covariance matrix $\Sigma^{k}$. 
The parameters $\Theta = \{ \alpha^{k}, \bm{\mu}^{k}, \Sigma^{k} \}_{k=1}^{K}$ of a GMM can be efficiently learned using the expectation-maximization (EM) algorithm~\cite{dempster_maximum_1977}. The two-step approach of the EM algorithm guarantees convergence to a local maximum of the likelihood function.

A defining property of GMMs is that each mixture component follows a multivariate normal distribution. While mixtures of such components can approximate highly complex joint distributions in principle, finite mixtures may smooth sharp marginal features or heavy-tailed behavior unless a sufficiently large number of components is used, consistent with characteristics observed to some extent in our scenario parameters. In many finite-sample settings, this smoothing can also be beneficial, as it mitigates overfitting and improves generalization.

\subsection{COPULAS}
Building on \cite{tewari_parametric_2011}, we describe how copulas facilitate the separation of joint distribution estimation into two tasks: 
\begin{enumerate}
    \item estimating the marginal distributions and
    \item estimating the dependencies among the random variables.
\end{enumerate}
This separation allows joint distributions to incorporate marginal distributions from various parametric and non-parametric families. The mathematical basis for copula-based joint distribution/density estimation is provided by Sklar’s Theorem \cite{sklar_fonctions_1959}.

\textit{Sklar's Theorem (1959)}: Any multivariate joint distribution $F(x_1, x_2, \dots x_d)$ of random variables $X_1, X_2, \dots X_d$ can be written in terms of their univariate marginal distribution functions $F_j(x_j)$ and a copula function, $C: U[0,1]^d \to [0,1]$, which describes the dependence structure between the variables, as shown in \eqref{eq:sklar}.
\begin{align}
\label{eq:sklar}
\begin{split}
F(x_1, x_2, \dots x_d) &= \mathrm{Pr} [ X_{1} \leq x_{1}, X_{2} \leq x_{2},\hdots  X_{d} \leq x_{d} ] \\
&= C(F_1(x_1), F_2(x_2), \dots F_d(x_d)) \\
&= C(u_{1}, u_{2}, \dots u_{d})
\end{split}
\end{align}
Note, that the marginal distribution values $F_j(x_j)$ are uniformly distributed in the interval $(0, 1)$ by definition and get abbreviated as $u_j$ in the following. As discussed in Tewari et al. \cite{tewari_parametric_2011}, the detailed proof of Sklar’s Theorem and its properties can be found in \cite{nelsen_introduction_1999, rank_estimation_2007, rank_coping_2007, clemen_correlations_1999}.

When both the copula function $C$ and the marginal distributions $F_j$ are differentiable, the joint density function can be expressed as the product of individual marginal densities $f_j$ and the copula density $c$ \cite{tewari_parametric_2011}:
\begin{align}
f(x_{1}, x_{2},\hdots x_{d}) &=  
f_{1}(x_{1}) \times f_{2}(x_{2}) \times \hdots \times f_{d}(x_{d}) \times \notag \\ 
&\quad c \left( u_1, u_2,\hdots u_d \right)
\label{eq:pdf}
\end{align}
This enables the estimation of joint densities while maintaining arbitrary marginal distributions and can be rearranged as \cite{tewari_parametric_2011}:
\begin{align}
    c(u_{1}, u_{2}, \hdots u_{d}) &= 
    \frac{f(x_1, x_2, \dots x_d)} {f_{1}(x_{1}) \times f_{2}(x_{2}) \times \hdots \times f_{d}(x_{d})}
    \label{eq:cdf}
\end{align}

Equation \eqref{eq:pdf} is primarily useful for estimating unknown joint densities given known marginal and copula densities, whereas \eqref{eq:cdf} allows deriving a copula density from a known multivariate joint density. A widely applied example is the Gaussian copula, derived from a multivariate Gaussian density, frequently used in finance and econometrics \cite{clemen_correlations_1999, li_default_1999}. However, their limitations in modeling multimodal datasets motivated \cite{tewari_parametric_2011} in developing a more flexible model, the GMCM.

\subsection{GAUSSIAN MIXTURE COPULA FUNCTION}
The Gaussian Mixture Copula (GMC) function is derived using the GMM density \eqref{eq:gmm} and \eqref{eq:cdf}, as described in \cite{tewari_parametric_2011}.
\begin{equation}
c_{gmc}(u_1, u_2, \dots u_d; \Theta) = \frac{\psi(z_1, z_2, \dots z_d; \Theta)}{\prod_{j=1}^{d} \psi_j(z_j)}
\label{eq:c_gmc}
\end{equation}
where, $z_1, z_2, \dots z_d$ are the inverse distribution values i.e. $z_j = \Psi_j^{-1}(u_j)$ and $\psi_j$ and $\Psi_j^{-1}$ denote the marginal density and inverse distribution function of the GMM along the $j^{th}$ dimension \cite{tewari_parametric_2011}.

\section{ESTIMATION OF GMCM PARAMETERS}
\label{sec:estimation}

As stated previously copulas facilitate the separation of joint distribution estimation from a training dataset $X \in \mathbb{R}^{d \times n}$ into two tasks, which also holds true for GMCMs: 
\begin{enumerate}
    \item estimating the marginal distributions $F_j(x_j)$
    \item estimating the dependencies $\Theta$ among the random variables 
\end{enumerate}

The marginal densities respectively distributions can be estimated using either parametric or non-parametric methods. Parametric estimation involves fitting predefined distributions such as a Lognormal or Beta distribution, while non-parametric methods, such as kernel density estimation (KDE), provide flexibility when the underlying distribution is unknown~\cite{parzen_estimation_1962, silverman_density_2018}.

Estimating the dependencies among the random variables, represented by the GMC parameters $\Theta = \left\{ \alpha^{k}, \< \bm{\mu}^{k}, \< \Sigma^{k} \right\}_{k=1}^{K}$ is non-trivial. It is neither possible to directly determine these from an empirical correlation matrix, as is the case for Gaussian copulas, nor is the EM algorithm for GMM parameter estimation directly applicable. The (log-)likelihood function \eqref{eq:ll} of GMCMs does not admit a closed-form analytical expression, in particular because of the function $\Psi^{-1}$ appearing in it. Further, GMCMs suffer from an inherent issue of parameter non-identifiability. These challenges and structural properties have been previously examined by Tewari et al. \cite{tewari_parametric_2011, tewari_estimation_2023}.
\begin{align}
    \ell_{c_{gmc}}(\Theta\mid U)
    = 
    \sum_{i=1}^{n} \log \frac{\psi\left( \Psi^{-1}(U_{:i}); \Theta \right)}{\prod_{j=1}^d \psi_j\left( \Psi^{-1}_j(U_{ji}; \Theta_{j})\right)}
    \label{eq:ll}
\end{align}

Thus, we estimated the GMC parameters using the maximum-a-priori (MAP) procedure, as proposed and implemented in \cite{tewari_estimation_2023}. The MAP estimates are calculated via numerical optimization, more precisely Stochastic Gradient Descent. Therefore, following \cite{tewari_estimation_2023}, we synthesize a GMCM as a transformed distribution using probabilistic programming languages. The base distribution (a GMM) undergoes two bijective mappings: first, marginal distribution functions $\Psi_j(\cdot)$ of the base GMM distribution, and second, quantile functions $F_j^{-1}(\cdot)$ of previously estimated marginal distributions. Consequently, the generative process of the GMCM can be specified as follows \cite{tewari_estimation_2023}:
\begin{subequations}
\begin{align}
\textbf{z} &\in \mathbb{R}^d \sim \mathrm{GMM}(\Theta)\notag\\
\textbf{u} &\in [0,1]^d
  = \bigl[\Psi_1(z_1;\Theta_1),\,\Psi_2(z_2;\Theta_2),\,\dots\,\Psi_d(z_d;\Theta_d)\bigr]
  \label{eq:4a}\\
\textbf{x} &\in \mathbb{V}^d
  = \bigl[F^{-1}_1(u_1),\,F^{-1}_2(u_2),\,\dots\,F^{-1}_d(u_d)\bigr],
  \label{eq:4b}
\end{align}
\end{subequations}
where the vector space $\mathbb{V}^{d}$ is formed by the Cartesian product of the supports of the $d$ marginal distribution functions $F_j$.

\begin{figure}
\centering\includegraphics[width=1.0\columnwidth]{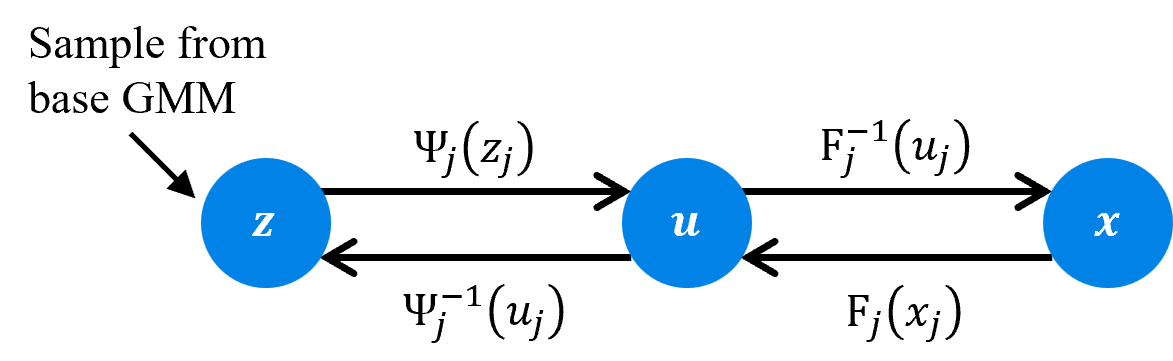}
\caption{Visualization of the transformation process in a GMCM. Starting from samples $\textbf{z}$ drawn from a GMM, two successive bijective transformations are applied independently along each dimension. The first maps each component through the cumulative distribution functions (CDFs) of the GMM marginals, followed by the application of inverse CDFs (quantile functions) of the marginal distributions previously estimated from data. Figure concept and caption adapted from Tewari et al. \cite{tewari_estimation_2023}.}
\label{fig:fig5}
\end{figure}

The generative process is also depicted in Figure \eqref{fig:fig5}. The joint density in the transformed space $X$ can be obtained by inverting a sample to the base space $Z$ and invoking the change of variable formula as shown in \eqref{eq:change_of_variable} \cite{tewari_estimation_2023}. The bracketed expressions correspond to the determinants of the Jacobians of the two transformations. These transformations have been implemented using the Bijector class available in TensorFlow Probability \cite{dillon_tensorflow_2017}.
\begin{align}
p(\textbf{x}) = \Biggl(\prod_{j=1}^d \frac{dF_j(x_j)}{dx_j}\Biggr)
\;\cdot\;
\Biggl(\prod_{j=1}^d \frac{d\Psi_j^{-1}(u_j)}{d u_j}\Biggr)
\;\cdot\;
\psi(\textbf{z})
\label{eq:change_of_variable}
\end{align}

To compute a univariate GMM’s quantile, $\Psi_j^{-1}(u_j;\Theta_j)$ we utilize the parallel implementation of Chandrupatla's algorithm \cite{chandrupatla_new_1997} via the Tensor Flow Probability package as suggested by \cite{tewari_estimation_2023}. The partial derivatives of $\Psi_j^{-1}\bigl(\,\cdot\,;\Theta_j\bigr)$ with respect to $\Theta_j = \{\alpha^{k},\,\mu_{j}^{k},\,\sigma_{j}^k\}_{k=1}^{K}$ can be obtained analytically despite the quantile function lacking a closed form. These derivations are provided in Appendix D of~\cite{tewari_estimation_2023}.

The identifiability issue in GMCs can be resolved by introducing Gaussian priors as shown in \eqref{eq:prior1} and \eqref{eq:prior2} \cite{tewari_estimation_2023}. The strength of these priors can be adjusted via the parameter $\sigma$, where larger values imply weaker priors.
\begin{equation}
\mathcal{N}\!\Bigl(\sum_{k=1}^{K}\alpha^{k}\, \mu_{j}^{k}\;\big\vert\;0,\sigma\Bigr),
\label{eq:prior1}
\end{equation}
\begin{equation}
\mathcal{N}\!\Bigl(\sum_{k=1}^{K}\bigl[\alpha^{k}\bigl(\Sigma^{k}_{jj}+(\mu_{j}^{k})^{2}\bigr)\bigr]\big\vert\;1,\sigma\Bigr).
\label{eq:prior2}
\end{equation}

\section{EXPERIMENT}
\label{sec:experiment} 
In this section, we present our experimental study of GMCMs for joint probability estimation of scenario parameters. Our experiments, conducted on two real-world datasets from certification-relevant scenarios, show that GMCMs consistently outperform GCMs. With comparable computational resources, GMCMs achieve higher accuracy than GMMs for one scenario and comparable performance for the other, indicating that both approaches are competitive and that performance depends on the specific data characteristics. These findings are consistent with previously reported results on GMCMs~\cite{tewari_parametric_2011, tewari_estimation_2023} and extend them to the domain of scenario parameter modeling.

\subsection{REAL-WORLD SCENARIO DATASETS}
\label{sec:datasets}
Our investigation was conducted on real-world datasets for two critical scenarios described in the UN R157:
    \begin{enumerate}
        \item \textit{Deceleration}: ``the ‘other vehicle’ suddenly decelerates in front of the ‘ego vehicle’''\cite[Annex~4, Appendix~3, para.~2.2(c)]{noauthor_uniform_2021}
        \item \textit{Follow Lead Vehicle Emergency Brake}
    \end{enumerate}

The scenario \textit{Deceleration} is provided as one of three traffic critical disturbance scenarios to define conditions under which ALKS shall avoid a collision. The scenario conditions (i.e., parameter values) are determined by simulating the scenarios with an attentive human-driver performance model.

For the description of the \textit{Deceleration} scenario, UN R157 separates between parameters for a) road geometry and b) other vehicle's behavior/ maneuver. In this work we focused on the parameters for other vehicle's behavior/ maneuver, like in the simulation examples provided by UN R157 and depicted in Fig. \ref{fig:deceleration}. The description of these parameters, analogous to UN R157, can be found in Table \ref{table:decel_params} below.

\setlength{\tabcolsep}{2pt}          
\renewcommand{\arraystretch}{1.0}

\begin{table}[h]
\caption{Parameters for \textit{Deceleration} scenario}
\label{table}
  \centering
  \begin{tabularx}{\columnwidth}{|l|l|l|X|}
    \hline
    \multirow[t]{3}{*}{Initial condition}
      & \multirow[t]{2}{*}{Initial velocity}
      & \textbf{Ve0}
      & Ego vehicle velocity (kph) \\ \cline{3-4}
    &
      & \textbf{Vo0}
      & Leading vehicle velocity (kph) \\ \cline{2-4}
    & Initial distance
      & \textbf{dx0}
      & Longitudinal distance (m) \\ \hline

    \multirow[t]{2}{*}{Vehicle motion}
      & \multirow[t]{2}{*}{Deceleration}
      & \textbf{Gx\_max}
      & Maximum deceleration (G) \\ \cline{3-4}
    &
      & \textbf{dG/dt}
      & Deceleration rate (Gps) \\ \hline
  \end{tabularx}
\label{table:decel_params}
\end{table}

UN R157 also contains test scenarios at a functional level, which are to be carried out in close coordination with a technical service. However, their definition itself leaves room for interpretation, therefore BMW published a common interpretation with partners, conducted in the context of the German research project SET Level~\cite{noauthor_sl-3-1-osc-alks-scenarios_nodate}. In detail, they implemented the test scenarios using international standards: OpenSCENARIO XML for scenario definitions and OpenDRIVE for road network definitions, which are executable with standard compliant simulators like esmini~\cite{noauthor_esmini_nodate}.

BMW provided 15 concrete parametrized test scenarios, each complemented by a variation file, which allows parameter variation~\cite{noauthor_sl-3-1-osc-alks-scenarios_nodate}. The scenario \textit{Follow Lead Vehicle Emergency Brake} is one of them, which is quite similar to the previously described \textit{Deceleration} scenario except from slight difference in parametrization. Also here we focus on continuous parameters describing the vehicles' dynamic behavior and therefore neglect the parameters \textbf{Road} and \textbf{Ego InitPosition LaneID}. Furthermore we do only consider scenarios with cars as \textbf{LeadVehicle Model}. The remaining parameters are shown in Table \ref{table:emerg_params}.

\setlength{\tabcolsep}{2pt}          
\renewcommand{\arraystretch}{1.0}

\begin{table}[h]
\caption{Parameters for \textit{Follow Lead Vehicle Emergency Brake} scenario}
\label{table:emerg_params}
  \centering
  \begin{tabularx}{\columnwidth}{|l|l|X|}
    \hline
    \multirow[t]{3}{*}{Initial condition}
      & Initial velocity
      & \textbf{Ego InitSpeed Ve0 (kph)} \\ \cline{2-3}
    & \multirow[t]{2}{*}{Initial distance}
      & \textbf{LeadVehicle Init HeadwayTime (s)} \\ \cline{3-3}
    & 
      & \textbf{LeadVehicle Init LateralOffset (m)} \\ \hline

    Vehicle motion
      & Deceleration
      & \textbf{LeadVehicle Deceleration Rate (mps2)} \\ \hline
  \end{tabularx}
\end{table}

For our investigation, we analyzed 13,613,530 and 4,111,855 real-world instances of the \textit{Follow Lead Vehicle Emergency Brake} and \textit{Deceleration} scenario, respectively. The data is derived from the Mercedes-Benz customer fleet, which collects information whenever its assistance systems detect a critical traffic situation, provided that customers have given their consent. In such cases, a full report describing the situation is sent to the backend for further analysis. For more details about the data used in this study, refer to~\cite{mercedes_data}.

As the available data includes only critical situations and thus represents a subset of all possible scenarios, the probability densities derived from it reflect this subset rather than the full distribution of real-world occurrences. However, this does not compromise the informative value of the study. First, for the validation of automated driving systems, only the probabilities of critical scenarios are relevant. These probabilities could be computed using the a priori likelihood of the collected situations. Second, the primary objective of this work is to evaluate various methods, particularly GMCs, for estimating the joint probability density of scenario parameters, rather than to derive parameter distributions for commercial validation. Consequently, the reduced scenario space in the datasets does not affect the validity of the study.

\subsection{RESULTS}

\begin{figure*}
  \centering
  \includegraphics[width=\textwidth]{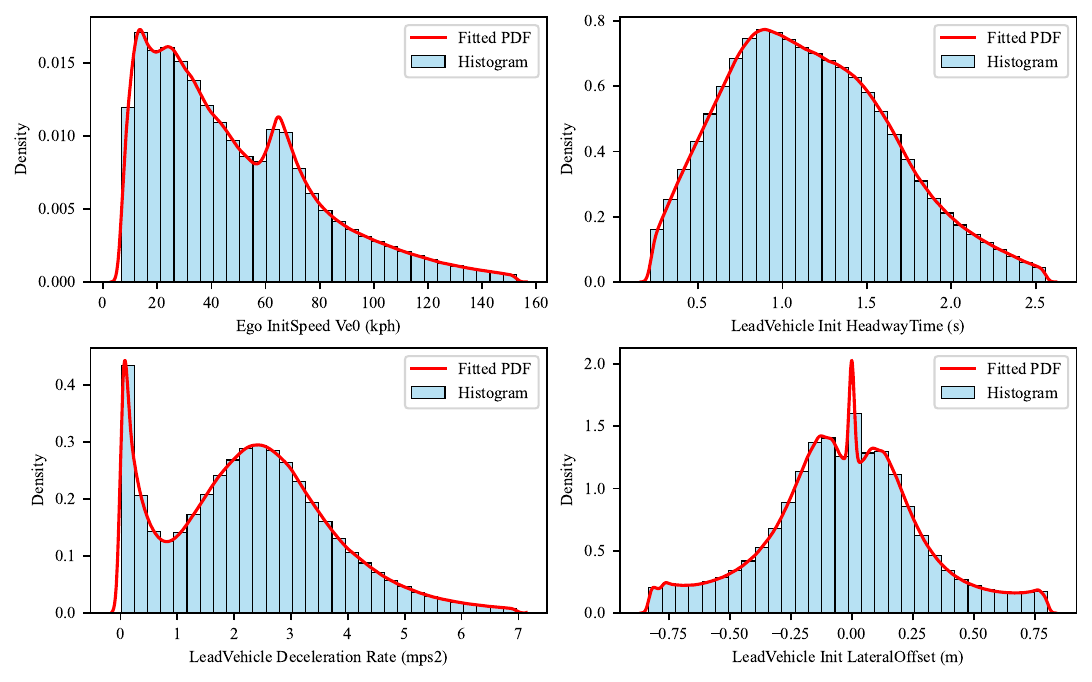}
  \caption{Marginal probability density functions (PDFs) for each parameter of the Follow‐Lead‐Vehicle Emergency Braking scenario, estimated via kernel density estimation. Real‐world data distributions are overlaid as histograms.}
  \label{fig:pdfs}
\end{figure*}

In the first step, we estimated the marginal distributions $F_j(x_j)$ of our real-world scenario datasets using KDE with Gaussian kernels, where the bandwidth was selected according to Scott’s rule \cite{scott_optimal_1979}. While parametric methods could also be employed (see Section \ref{sec:estimation}), we opted for a non-parametric approach due to the presence of multimodality and non-Gaussian components in the data. KDE offers the flexibility to capture such complex distributions more effectively. Fig. \ref{fig:pdfs} shows the estimated densities obtained using this procedure for the Follow Lead Emergency Brake scenario, along with the original marginal distributions represented as histograms.

In the second step, the MAP procedure for the estimation of the dependencies $\Theta$ among the scenario parameters was carried out using the Adam optimizer with the default learning rate of $10^{-3}$ \cite{kingma_adam_2014}. The GMCMs were trained on both datasets with four mixture components. An automated selection using criteria such as the Bayesian Information Criterion (BIC) is proposed for future work (see Section~\ref{sec:conclusion}).

To ensure a fair comparison, we matched the GMM component count to a comparable compute-time budget measured on the same machine. Specifically, we ran the four-component GMCM five times and used the median wall-clock time as the budget. We then chose the GMM component count so that its median wall-clock time (over five runs) matched this budget most closely. This procedure was applied independently to each dataset, resulting in GMMs with twelve components for the \textit{Follow Lead Vehicle Emergency Brake} dataset and ten components for the \textit{Deceleration} dataset. All experiments were run on a machine with an Intel Core i7-1365U (10~cores/12~threads) and 32~GB RAM running Windows~11. No GPU acceleration was used.

\begin{figure*}
  \centering
  \includegraphics[width=\textwidth]{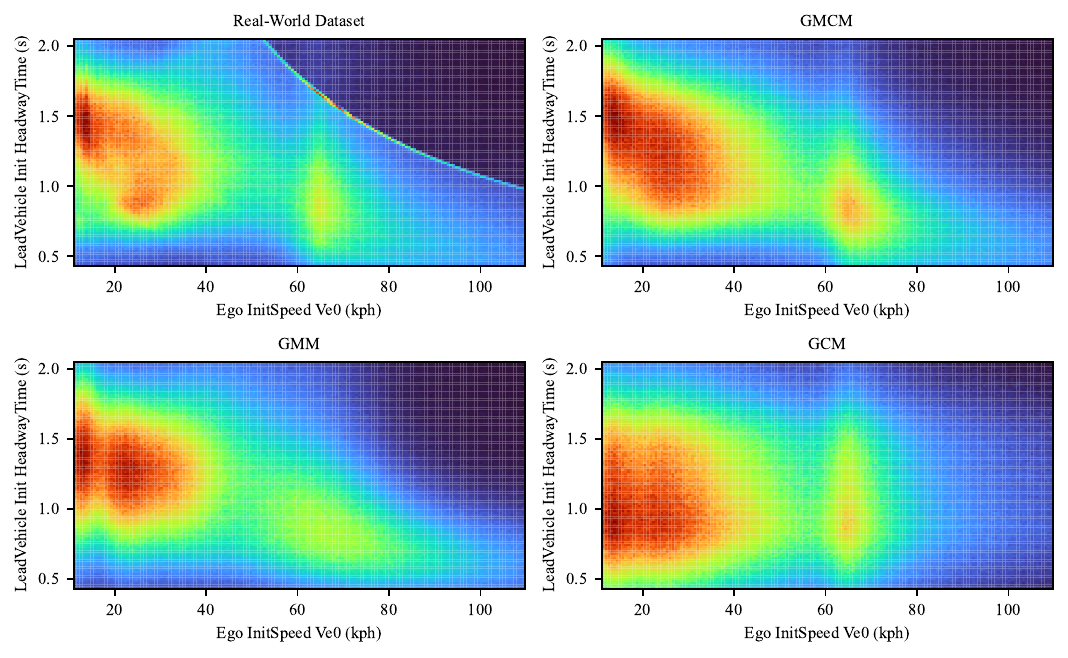}
  \caption{Bivariate  marginal distribution of the original data and samples from GMCM, GMM, and GCM, illustrated via density heatmaps (colour scale: turbo).}
  \label{fig:heatmap}
\end{figure*}

To illustrate the achieved expressive power of the GMCM, we visualize the bivariate marginal distribution of \textbf{LeadVehicle Init HeadwayTime (s)} and \textbf{Ego InitSpeed Ve0 (kph)} from the \textit{Follow Lead Vehicle Emergency Brake} dataset via density heatmaps. Fig. \ref{fig:heatmap} compares the original data distribution—which appears multimodal with non-Gaussian modes—with the estimated densities from the GMCM, the GMM, and the GCM. 

In the high-density region on the left, all three models deviate from the empirical distribution, but both GMCM and GMM still capture the main structure with a descending trend of \textbf{LeadVehicle Init HeadwayTime (s)} towards higher \textbf{Ego InitSpeed Ve0 (kph)}, whereas the GCM tends to smooth out this pattern. In the central high-density region and the right part of the plot, the GMCM more faithfully reproduces the location and shape of the empirical distribution. In contrast, the GMM smooths out the central high-density region, and the GCM assigns too much density to the upper-right part of the plot. Overall, the visual comparison of the bivariate density heatmaps suggests that the GMCM provides the closest match to the real data distribution.

Unlike the heatmaps, which only allow model evaluation in two dimensions, this observation is quantitatively supported across the full-dimensional space by the Sinkhorn distance and the log-likelihood, where the GMCM demonstrates improved performance—most clearly in the Sinkhorn distance—while achieving a log-likelihood comparable to the GMM. In an analogous evaluation on the \textit{Deceleration} scenario dataset using the same metrics, the GMCM performs comparably to the GMM, with the GMM holding a slight edge. The GCM performs markedly worse, as shown in Table~\ref{tab:results}.

\begin{table}
\caption{Log-likelihood, Sinkhorn distance, and median training wall-clock time achieved by GMCM, GCM, and GMM on our \textit{Follow Lead Vehicle Emergency Brake} and \textit{Deceleration} datasets. Labels of the form GMM-K and GMCM-K indicate the number of mixture components K. The best-performing model for each metric and dataset is shown in \textbf{bold}.}
\centering
\setlength{\tabcolsep}{6pt}

\begin{minipage}{\linewidth}
\centering
\begin{tabular}{@{}p{0.135\linewidth} p{0.215\linewidth} p{0.26\linewidth} p{0.24\linewidth}@{}}
\toprule
\textbf{Model} & \textbf{Log-Likelihood} & \textbf{Sinkhorn Distance} & \textbf{Wall-Clock Time} \\
\midrule
GMM-12 & -6.8610 & 0.542 & 1546 s\\
GCM  & -7.1706 & 0.511 & 41 s\\
GMCM-4 & \textbf{-6.7923} & \textbf{0.436} & 1707 s\\
\bottomrule
\end{tabular}
\vspace{0.5em}

(a) \textit{Follow Lead Vehicle Emergency Brake}
\end{minipage}

\vspace{1em}

\begin{minipage}{\linewidth}
\centering
\begin{tabular}{@{}p{0.135\linewidth} p{0.215\linewidth} p{0.26\linewidth} p{0.24\linewidth}@{}}
\toprule
\textbf{Model} & \textbf{Log-Likelihood} & \textbf{Sinkhorn Distance} & \textbf{Wall-Clock Time} \\
\midrule
GMM-10 & \textbf{-9.3703} & \textbf{1.65} & 361 s\\
GCM  & -12.0305 & 11.8 & 15 s\\
GMCM-4 & -9.6185 & 1.69 & 289 s\\
\bottomrule
\end{tabular}
\vspace{0.5em}

(b) \textit{Deceleration}
\end{minipage}

\label{tab:results}
\end{table}

The Sinkhorn distance is an entropically-regularized approximation of the classic Wasserstein (a.k.a. Earth Mover’s) distance between two probability distributions. Intuitively, the Sinkhorn distance measures how “far” two sample sets are by computing the least cost of transporting the mass of one distribution to match the other, while allowing some “fuzziness” (entropy) in the matching. In practice, a smaller Sinkhorn distance indicates that the generated samples closely mimic the target data distribution in terms of its geometry and support. For more details, refer to \cite{cuturi_sinkhorn_2013}.

Because computing Sinkhorn distances over very large datasets remains computationally demanding, we computed the distance as the mean over ten independent subsets of 100,000 samples each. For this purpose, we used the open-source Sinkhorn implementation \cite{williams_scalable-pytorch-sinkhorn_nodate}. The observed standard deviations within each model were small, and crucially, there was no overlap between the models’ confidence intervals. This confirms that, despite subsampling, Sinkhorn distances yield reliable relative comparisons.

The Python implementation we used to achieve the results presented in this section is provided with the paper~\cite{sourcecode}.

\section{CONCLUSION}
\label{sec:conclusion}

This paper presents the first investigation of GMCMs for estimating joint distributions of scenario parameters in the context of scenario-based validation for automated driving systems. We compared GMCMs with previously proposed models for this application—GMMs and GCMs. This comparison confirmed that the advantages reported in earlier studies by Tewari et al. \cite{tewari_parametric_2011, tewari_estimation_2023} generalize effectively to the domain of scenario parameter modeling for automated driving. GMCMs combine the ability of GMMs to represent multimodal distributions with the flexibility of copulas to model arbitrary marginal distributions independently of their dependence structure.

Our evaluation was based on two large-scale real-world datasets comprising approximately 14 million and 4 million driving scenario instances, respectively. The scenario types were selected based on the approval regulation UN R157 for Society of Automotive Engineers (SAE) Level 3 conditional driving automation functions~\cite{on-road_automated_driving_orad_committee_taxonomy_nodate}. To the best of our knowledge, neither GMMs nor GCMs have been evaluated on scenario datasets of comparable size and degree of reality. We assessed model performance using three complementary metrics: visual plausibility through heatmaps, log-likelihood, and Sinkhorn distance. To promote reproducibility and further research, we also provide a Python implementation that enables the training, evaluation, and application of GMCMs on arbitrary datasets.

GMCMs demonstrated improved performance over GMMs in one dataset and comparable performance (with a slight GMM edge) in the other, while consistently outperforming GCMs, even without hyperparameter fine-tuning. Importantly, GMCMs retain the dependence structure of GMMs. Their benefits arise from decoupling marginal and dependence modeling, which can be advantageous when accurate marginal modeling is particularly important, for example in the presence of heavy tails or sharp, localized marginal features. This suggests that GMCMs offer a complementary modeling alternative that emphasizes flexible marginal estimation, leading to a different bias--variance trade-off compared to standard GMMs. 

Future work should explore the extent to which performance can be improved by tuning estimation parameters, such as selecting the optimal number of mixture components via the BIC and adjusting convergence criteria~\cite{schwarz_estimating_1978}. Moreover, GMCMs should be applied to other scenario types defined in UN R157, such as Cut-in and Cut-out. Their performance in higher-dimensional settings should also be evaluated, for example by incorporating parameters related to roadway and environmental conditions. Finally, we restricted our analysis to continuous parameters. How discrete scenario parameters could be integrated into the GMCM framework remains an open question and is left for future work.

In summary, GMCMs offer a powerful and flexible framework for the statistical modeling of complex driving scenarios and show great promise for advancing scenario-based safety assessments in automated driving.

\bibliographystyle{IEEEtran}
\bibliography{references}

\end{document}